\documentclass[letterpaper]{article} 
\usepackage{aaai24}  
\usepackage{times}  
\usepackage{helvet} 
\usepackage{courier}  
\usepackage[hyphens]{url}  
\usepackage{graphicx}
\usepackage{natbib}
\usepackage{caption} 
\usepackage[ruled,linesnumbered]{algorithm2e}
\usepackage{diagbox}
\usepackage{amsmath}  
\usepackage{amssymb}  
\usepackage{booktabs} 
\usepackage{multirow} 
\usepackage{array} 
\usepackage{subcaption}

\usepackage{xcolor}
\usepackage{appendix}

\title{PG-LBO: Enhancing High-Dimensional Bayesian Optimization with Pseudo-Label and Gaussian Process Guidance}
\author {
    Taicai Chen\textsuperscript{\rm 1},
    Yue Duan\textsuperscript{\rm 1},
    Dong Li\textsuperscript{\rm 2},
    Lei Qi\textsuperscript{\rm 3},
    Yinghuan Shi\textsuperscript{\rm 1,}\thanks{Corresponding author. This work is supported by the National Key R\&D Program of China (2022ZD0116408), the Science and Technology Innovation 2030 New Generation Artificial Intelligence Major Projects (SQ2023AAA010051), NSFC Program (62222604, 62206052, 62192783), Jiangsu Natural Science Foundation Project (BK20210224), Huawei-Nanjing University Technical Cooperation Project (20221108058).},
    Yang Gao\textsuperscript{\rm 1}
}
\affiliations {
    \textsuperscript{\rm 1}National Key Laboratory for Novel Software Technology, Nanjing University, China\\
    \textsuperscript{\rm 2}Huawei Noah's Ark Lab, China\\
    \textsuperscript{\rm 3}School of Computer Science and Engineering, Southeast University, China\\
    \{taicaichen, yueduan\}@smail.nju.edu.cn, lidong106@huawei.com, qilei@seu.edu.cn, \{syh, gaoy\}@nju.edu.cn 
}
\usepackage{bibentry}

\begin{document}
\maketitle
\begin{abstract}
Variational Autoencoder based Bayesian Optimization (VAE-BO) has demonstrated its excellent performance in addressing high-dimensional structured optimization problems. However, current mainstream methods overlook the potential of utilizing a pool of unlabeled data to construct the latent space, while only concentrating on designing sophisticated models to leverage the labeled data. Despite their effective usage of labeled data, these methods often require extra network structures, additional procedure, resulting in computational inefficiency. To address this issue, we propose a novel method to effectively utilize unlabeled data with the guidance of labeled data. Specifically, we tailor the pseudo-labeling technique from semi-supervised learning to explicitly reveal the relative magnitudes of optimization objective values hidden within the unlabeled data. Based on this technique, we assign appropriate training weights to unlabeled data to enhance the construction of a discriminative latent space. Furthermore, we treat the VAE encoder and the Gaussian Process (GP) in Bayesian optimization as a unified deep kernel learning process, allowing the direct utilization of labeled data, which we term as Gaussian Process guidance. This directly and effectively integrates the goal of improving GP accuracy into the VAE training, thereby guiding the construction of the latent space. The extensive experiments demonstrate that our proposed method outperforms existing VAE-BO algorithms in various optimization scenarios. Our code will be published at \textcolor{magenta}{https://github.com/TaicaiChen/PG-LBO}.
\end{abstract}

\section{Introduction}
Bayesian optimization has found widespread applications in various optimization problems and machine learning tasks, enabling the discovery of optimal solutions for complex functions \cite{snoek2012practical}. Traditional BO methods based on the Gaussian process have demonstrated their remarkable performance in practical applications. However, with the increase in problem dimensions, these conventional methods face significant challenges. The computational cost of searching in high-dimensional spaces grows exponentially, leading to a drastic decrease in the efficiency of traditional BO algorithms \cite{kandasamy2015high}. Recently, Variational Autoencoder (VAE) provides a powerful method to map high-dimensional structured data into low-dimensional continuous latent spaces, enabling efficient and effective optimization of the objective function. Consequently, VAE-BO possesses unique advantages in tackling high-dimensional structured optimization problems. Currently, the VAE-BO based methods have been widely applied in various fields, including molecular synthesis \cite{stanton2022accelerating, maus2023inverse}, neural architecture search \cite{ru2021interpretable, biswas2023optimizing} and anomaly detection \cite{zhang2023hybrid}. The research focus of VAE-BO lies in effectively utilizing both labeled and unlabeled data to learn a discriminative latent space. Nevertheless, some existing VAE-BO methods still exhibit limitations in the utilization of labeled data and unlabeled data.

Firstly, while existing VAE-BO algorithms make use of labeled data, unfortunately, their utilization of unlabeled data is still straightforward, whereas the efficacy of unlabeled data has not been fully exploited. For example, the pioneering work of \cite{gomez2018automatic} was the first to introduce VAE into the BO domain. However, in their method, the VAE was trained in an unsupervised manner, resulting in a non-discriminative latent space that was ill-suited to BO processes. Subsequent VAE-BO research has largely focused on better utilizing labeled data with black-box function values to learn a discriminative latent space. For instance, \cite{eissman2018bayesian} introduced additional network structures that allow for the joint utilization of labeled and unlabeled data in the pre-training of the VAE. While LBO \cite{tripp2020sample} employed unlabeled data for the initial pre-training of the VAE and subsequently leveraged labeled data through data weighting and periodic retraining of the VAE. In addition, T-LBO \cite{grosnit2021high} extended LBO by incorporating deep metric learning to utilize labeled data, and the training of the VAE was divided into two stages: pre-training on unlabeled data and fine-tuning on labeled data. These works proposed various ways to leverage labeled data, while the utilization of unlabeled data has only been explored in the pre-training stage of the VAE, and its potential has not been fully exploited. We notice that unlabeled data also contain valuable discriminative information regarding optimization target values. Exploiting this information can help train more discriminative models and learn a better latent space.

Furthermore, we have observed that existing methods face challenges of complexity or inefficiency when it comes to utilizing labeled data. \cite{eissman2018bayesian} required to introduce additional network structures which predicted black-box function values through the encoded latent representations, making the method somewhat complex. And T-LBO introduced significant computational overhead due to deep metric learning, and the overall method was computationally demanding. LBO utilized the relative magnitudes of labeled data labels through data weighting, rather than directly using the actual values, potentially not fully harnessing the potential of labeled data. Hence, how to effectively utilize labeled data in a simple yet efficient manner is worth investigating.

Be aware of these two issues, we wonder if we have an effective and efficient way to  fully capitalize the potential of both labeled and unlabeled data. To this end, we present a novel semi-supervised learning method that utilizes pseudo-labels to explicitly reveal the relative magnitudes of optimization target scores in unlabeled data. Subsequently, we assign appropriate training weights to unlabeled data based on these pseudo-labels. By assigning more weights to the points with higher optimization target scores, this guides VAE to focus more on high-scoring points. This contributes to the construction of a more discriminative latent space. Furthermore, we propose a more streamlined method for exploiting labeled data. Specifically, we unify the VAE encoder and the GP in BO as a cohesive deep kernel learning process, effectively integrating the goal of improving GP accuracy directly into the VAE training. This integration allows the GP to guide the training of the VAE and consequently influence the construction of the latent space. The main contributions of our work can be summarized as follows:
\begin{itemize}
    \item We assess the limitations of current VAE-BO approaches in terms of data utilization and delve into the untapped potential of unlabeled data in enhancing the creation of a distinct latent space.
    \item For the comprehensive usage of unlabeled data, we propose to utilize pseudo-labels to intergrate the implicit discriminative information within unlabeled data, thereby facilitating the construction of the latent space.
    \item To efficiently exploit the labeled data, we introduce a straightforward method through GP guidance by integrating the VAE encoder and GP to leverage labeled data, modeled as a classical deep kernel learning paradigm.
\end{itemize}

\begin{figure}[t]
\centering
\includegraphics[width=0.9\columnwidth]{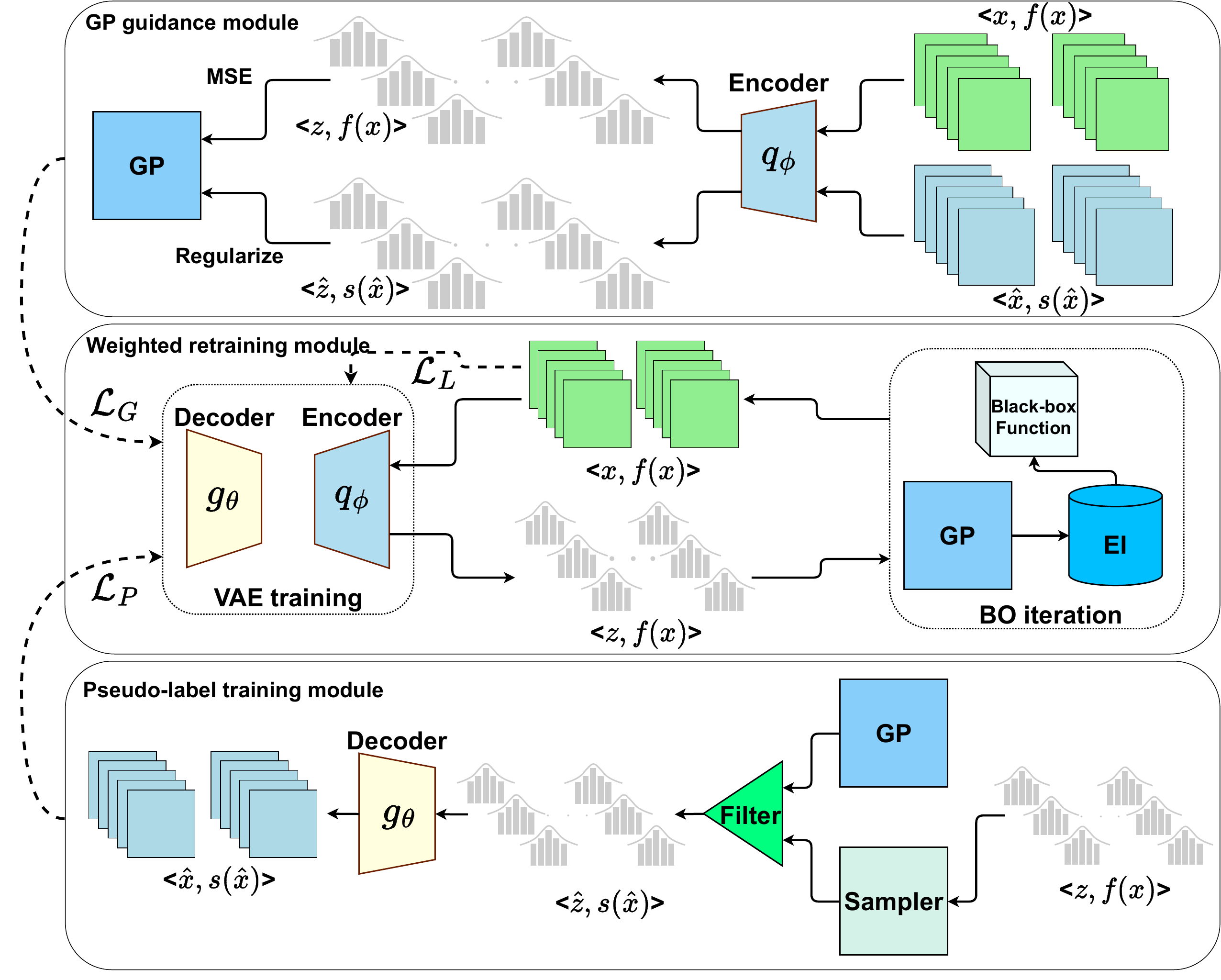}
\caption{A overview of the PG-LBO.}
\label{fig:overview}
\end{figure}

\section{Related Work}
Bayesian optimization suffered from the curse of dimensionality, which referred to the exponential increase in computational complexity as the dimension of the problem increased. Although various methods had been proposed to extend Bayesian optimization to high-dimensional spaces, the focus in this paper was on the VAE-based method, also known as Latent Space Optimization (LSO) \cite{tripp2020sample}. The VAE was used to map the high-dimensional structured space $\mathcal{X}$ to a lower-dimensional continuous latent space $\mathcal{Z}$. In the latent space $\mathcal{Z}$, Bayesian optimization was performed to find the optimal solution $z^\star \in \mathcal{Z}$, which was subsequently mapped back to $\mathcal{X}$ to obtain the final solution $x^\star$. A key challenge of VAE-BO lays in effectively leveraging labeled and unlabeled data.

The VAE-BO was first applied to chemical design \cite{gomez2018automatic} and found application in various domains including automatic machine learning \cite{zhang2019d, biswas2023optimizing}, chemical molecule synthesis \cite{korovina2020chembo, maus2023inverse} and anomaly detection \cite{zhang2023hybrid}. In those works, the VAE models were mostly trained on unlabeled data, and their parameters remained fixed during the BO process. This conventional paradigm led to a non-discriminative potential space ill-suited for BO and vulnerable to the ``dead zone" issue. To address these issues, LBO suggested integrating LSO with weighted retraining of the VAE model, leveraging the black-box function values of labeled data. Assigning weights to data points according to the magnitude of the black-box function scores compelled the generative model to prioritize modeling the feasible region with high-scoring points. Furthermore, periodic VAE model retraining during Bayesian optimization iterations facilitated the incorporation of new point information. For guiding the creation of a distinctive VAE-BO latent space, T-LBO extended LBO by incorporating deep metric learning to better exploit labeled data. In addition, T-LBO adopted a pre-training and fine-tuning framework, using unlabeled data for pre-training the VAE model and then supervised fine-tuning with labeled data. This training framework effectively combined labeled and unlabeled data. Recent work, TSBO \cite{yin2023high}, drew inspiration from Meta Pseudo Labels \cite{pham2021meta} in semi-supervised learning, introduced an additional teacher-student model structure to assign pseudo-label to the unlabeled data, which were then utilized for training the surrogate model and the VAE. 

Certain VAE-BO research redirected its attention to areas beyond data utilization. For example, LOL-BO \cite{maus2022local} concentrated on optimizing strategies within VAE-BO, introducing trust region optimization strategies that aided in searching for optimal points during the BO process. On the other hand, \cite{verma2022high} utilized Gaussian Process marginal likelihood objectives to learn invariant enhancements for determining BO query points. \cite{guo2022generative} involved using multiple VAEs to mitigate the challenges posed by confirming VAE latent space dimensions.

Our method emphasized the utilization of data in VAE-BO, specifically focusing on the utilization of unlabeled data. Compared with previous methods, we used pseudo-labels to exploit the implicit discriminative information in unlabeled data to help construct the VAE latent space, instead of simply using unlabeled data for VAE pre-training. TSBO also employed pseudo-labeling, but our method differed in pseudo-label’s application. TSBO directly utilized pseudo-labels to characterize black-box function values. While considering the potential cognitive errors introduced by pseudo-labels, we used pseudo-labels to reveal the magnitude relationship information between the black-box function values corresponding to unlabeled data and utilized this information using data weighting. Meanwhile, TSBO focused on using unlabeled data to help train the surrogate models, while we focused on using unlabeled data to aid in constructing the latent space of VAE.

\section{Methodology}
\subsection{Preliminaries: Bayesian Optimization}
Bayesian optimization addresses the global optimization problem of finding
\begin{align}
x^\star = \underset{x \in \mathcal{X}}{\text{argmax}}\, f(x),
\end{align}
where $f(\cdot):\mathcal{X} \to \mathbb{R}$ is an expensive black-box function defined on a high-dimensional structured input space $\mathcal{X}$. BO leverages two core components: a surrogate model to model the objective function and an acquisition function to determine the next query point for evaluation. Gaussian Process Regression (GP) is commonly used as the surrogate model, which models the posterior distribution $f_\Phi(x; D_L)$ based on the available data set $D_L=\{(x_i,y_i)|i=1,\cdots, N\}$, where $\Phi$ represents the hyperparameters of the GP and $y_i=f(x_i)$.

The posterior distribution is given as follows:
\begin{align}
f_\Phi(x;D_l) \sim \mathcal{N}(\mu(x),\sigma^2(x)),
\end{align}
where
\begin{align}
\mu(x) &= m(x) + k(x,X)^\top(K+\sigma_0^2I_N)^{-1}(y-m(x)), \\
\sigma^2(x) &= k(x,x) - k(x,X)^\top(K+\sigma_0^2I_N)^{-1}k(x,X),
\end{align}
where $m(\cdot)$ and $k(\cdot, \cdot)$ are the prior mean and kernel function, $I_N$ denotes the $N$-dimensional identity matrix, $k(x,X)=[k(x,x_1),k(x,x_2),\cdots,k(x,x_N)]$, and the elements of matrix $K$ are defined as $K_{ij}=k(x_i,x_j)$. GP is a non-parametric model, and its hyperparameters $\Phi$ mainly include the prior mean function, kernel function scale, and noise variance, which are typically optimized by minimizing the negative log marginal likelihood on the available data set.

The acquisition function is designed to balance the posterior mean and variance predicted by the GP model to determine the next query point. Popular acquisition functions include Expected Improvement (EI), Probability Improvement (PI), and Upper Confidence Bound (UCB), among others.

\subsection{Overview of the PG-LBO}
\subsubsection{Weighted retraining module.}
 Weighted retraining, proposed by LBO, employs data weighting to enhance the modeling of high-scoring points in the VAE training objective, allocating a larger proportion of the feasible region and effectively utilizing all known data points in order to learn informative representations and avoid overfitting. Let $q^{enc}_\phi$ be the pre-trained VAE encoder, $g^{dec}_\theta$ be the decoder. The initial available labeled data set is denoted as $D_L=\{(x_i,f(x_i))|i=1,2,\cdots,N\}$.

The widely used weighting method is based on a rank-based weighting function:
\begin{align}
w(x;D_L,k) &\propto \frac{1}{kN+\text{rank}_{f,D_L}(x)}, \\
\text{rank}_{f,D_L}(x)&=|\{x_i: f(x_i)>f(x), x_i \in D_L\}|.
\end{align}
This function assigns weights approximately proportional to the inverse rank (starting from zero) of each data point. A tunable hyperparameter $k$ controls the degree of weighting, where $k=\infty$ corresponds to uniform weighting ($w_i = \frac{1}{N}$ for all $i$), and $k=0$ assigns all weight to the single point with the highest objective function value. The VAE is trained on the weighted data set, and training loss can be seen as the weighted average over the data points:
\begin{align}
\nonumber \mathcal{L}_L(\theta ,\phi) &= w(x_i)[E _{q_{\phi}(z_i|x_i)}[\text{log}\,g_{\theta}(x_i|z_i)] 
\\ &-\text{KL}(q_{\phi}(z_i|x_i)||p(z))],
\end{align}
where $w(x_i)$ represents the data weight calculated according to the weighting function. 

In addition to data weighting, periodic retraining of the VAE is also necessary to propagate the newly generated points by the BO process. Specifically, assuming that retraining is performed every $r$ BO iterations, before each retraining, the labeled data set is updated as follows: $D^{(k)}_L = D^{(k-1)}_L \cup \{(x^\star_i,f(x^\star_i))\}_{i=0,...,q}$, and where $D^{(0)}_L = D_L$. New points are added to the labeled data set and then the updated data set is used to retrain the VAE.

\subsubsection{Pseudo-label training module.}
Existing VAE-BO methods primarily focus on utilizing the black-box function labels of labeled data to guide the construction of the latent space, while overlooking the potential role of implicit discriminative information regarding optimization target values in unlabeled data. In response, we introduce pseudo-label training, a novel approach leveraging pseudo-labels to uncover latent discriminative information within unlabeled data. This knowledge is effectively employed via weighted retraining to guide latent space construction. The training objective of VAE-BO can be seen as training a Gaussian process latent variable model \cite{siivola2021good}, resembling a deep regression task. However, unlike extensively studied classification tasks in semi-supervised learning that have explored settings such as mismatch \cite{duan2023towards}, open-set \cite{li2023iomatch,YangWSLZXY22} and barely-supervised learning \cite{gui2022improving}, the exploration of semi-supervised methodologies for deep regression tasks is relatively limited. Regression tasks directly use real-valued targets as pseudo-labels, being sensitive to prediction quality. However, Weighted retraining mitigates this sensitivity by indirectly incorporating relative objective function ranking through data weighting, avoiding direct utilization of actual values. This method reduces training sensitivity to black-box function values, enabling the effective use of unlabeled data via pseudo-labels.

Specifically, we sample unlabeled data points in the latent space $\mathcal{Z}$ to obtain a set of points $\{\hat{z}_j\}_{j=0,...,N_P}$, then assign pseudo-labels to these unlabeled data points to reflect the relative magnitudes of the black-box function values, resulting in a pseudo-labeled data set $D_P=\{\hat{x}_j, s(\hat{x}_j)|j=0,\cdots, N_P\}$, where $\hat{x}_j=g_\theta(\hat{z}_j)$. We then use the pseudo-labels to weigh the data during training. By introducing an additional pseudo-label loss $\mathcal{L}_P$ during training, it becomes integrated into the weighted retraining process of VAE:
\begin{align}
\nonumber \mathcal{L}_{P}(\theta ,\phi )&=\hat{w}(\hat{x}_j)[E _{q_{\phi}(\hat{z}_j|\hat{x}_j)}[\text{log}\,g_{\theta}(\hat{x}_j|\hat{z}_j)]
\\ &-\text{KL}(q_{\phi}(\hat{z}_j|\hat{x}_j)||p(z))], 
\\ \hat{w}(\hat{x};D_P,k) &\propto \frac{1}{kN_P+\text{rank}_{s,D_P}(\hat{x})},
\\ \text{rank}_{s,D_P}(\hat{x})&=|\{\hat{x}_j: s(\hat{x}_j)>s(\hat{x}), \hat{x}_j \in D_P\}|,
\end{align}
where $s(\hat{x})$ represents the pseudo-label of the unlabeled data sample $\hat{x}$, which only needs to reflect the relative magnitudes of the black-box function values among unlabeled data, rather than accurately reflecting the actual values. For simplicity, we directly use the posterior mean provided by the GP model as the pseudo-label, \textit{i.e.}, $s(\hat{x}_j)=\mu_{f_\Phi}(\hat{z}_j)$, where $\hat{x}_j = g_\theta(\hat{z}_j)$.

\subsubsection{GP guidance module.}
The core purpose of utilizing labeled data is to construct a more discriminative latent space tailored for the Bayesian optimization process. Drawing inspiration from the principles of deep kernel learning, we present a more direct approach called GP guidance to harness the black-box function values of labeled data. We treat the VAE encoder and the GP model as integral components of a unified deep kernel learning process. This technique introduces a loss term that aims to minimize the mean squared error between the GP model's predictions and the actual black-box function values of labeled data. This loss term is seamlessly integrated into the VAE's training objective, driving the VAE to generate latent representations that closely align with the GP model's expectations. Notably, the GP model offers a predictive posterior distribution that gauges the uncertainty of predictions. To enhance stability, we regularize the posterior by incorporating an unsupervised loss term designed to minimize the prediction variance for unlabeled data points \cite{jean2018semi}. The ultimate formulation of this loss term can be expressed as follows:
\begin{align}
\nonumber \mathcal{L}_G(\phi)&=w(x_i)E_{q_{\phi}(z_i|x_i)}[(f(x_i) - \mu_{f_\Phi}(z_i))^2]
\\&+\hat{w}(\hat{x}_j)E_{q_{\phi}(\hat{z}_j|\hat{x}_j)}[\sigma^2_{f_\Phi}(\hat{z}_j)].    
\end{align}
Since the GP model is a non-parametric model and lacks learnable parameters like neural networks, it primarily relies on the hyperparameters $\Phi$ to fit the training data. Therefore, during VAE training, we keep the GP model fixed and only use it for guidance. After VAE training, we re-encode the labeled data using VAE to obtain new data latent representations, then fit the GP model on these new representations, and update the hyperparameters of the GP model. This way, the VAE and the GP model are continually optimized through this approximate alternating update process.

\subsubsection{Data sampling and pseudo-label selection.}
Unlabeled data is acquired through sampling. Weighted retraining encourages the VAE to use larger feasible regions for modeling high-scoring points. However, if the sampled unlabeled data mainly comprises low-scoring points, it might not contribute effectively to the model and could even worsen its performance. Hence, the data sampling strategy must be carefully designed to prioritize higher-scoring points. In this study, we utilize two straightforward sampling approaches: noisy sampling and heuristic sampling. Noisy sampling introduces Gaussian noise to existing labeled data, effectively sampling around high-scoring points. Heuristic sampling treats sampling as an optimization task, utilizing a simple heuristic optimization algorithm to iteratively maximize data point pseudo-labels. These approaches enhance the overall quality of sampled points. A detailed introduction of the sampling methods are in Appendix A.

While data weighting effectively mitigates potential impacts from discrepancies between pseudo-labels and true labels, pseudo-labels carrying significant errors can still introduce cognitive bias into the model. Therefore, we filter the sampled data during the sampling process to improve the accuracy of the pseudo-label data as much as possible. Bayesian optimization can be viewed as an optimization for regression tasks. Different from classification tasks where straightforward thresholding can sift confident class predictions, regression tasks employ real-valued targets as pseudo-labels, rendering the sole reliance on model outputs for confidence assessment challenging. Inspired by the work of \cite{rizve2021in}, we use uncertainty to select robust pseudo-label data. This choice presents distinct benefits when coupled with Gaussian process models, as GP models can jointly express both mean and variance. The variance conveniently doubles as a measure of prediction uncertainty, negating the need for extra computation. We provide a more detailed discussion on the threshold selection in Appendix B. Drawing inspiration from the work of FreeMatch \cite{wang2023freematch} in semi-supervised learning, we set the uncertainty threshold $\tau_t$ for filtering as the model's average prediction variance on unlabeled data and estimate the uncertainty threshold as the exponential moving average (EMA) of threshold values at each training step:
\begin{align}
\tau_t = 
\begin{cases} 
\frac{1}{N^\prime_P}\sum^{N^\prime_P}_{j=1}\sigma^2_{f_\Phi}(z^\prime_j),  & \mbox{if }t\mbox{ = 0},\\
\lambda \tau_{t-1}+(1-\lambda)\frac{1}{N_P}\sum^{N_P}_{j=1}\sigma^2_{f_\Phi}(z_j), & \mbox{otherwise},
\end{cases}
\end{align}
where $\lambda \in (0,1)$ is the momentum decay for EMA. When $t=0$, we pre-sample a subset of points $\{z^\prime_j\}_{j=1,...,N^\prime_P}$, where $N^\prime_P = N_P/10$, to determine the initial threshold.

\subsubsection{Training loss.} By combining the aforementioned components, we have developed a novel high-dimensional Bayesian optimization algorithm, named PG-LBO. Figure \ref{fig:overview} provides an overview of the PG-LBO. The training loss of the algorithm can be represented as follows:
\begin{align}
\mathcal{L}=\mathcal{L}_{L} + \lambda_P \mathcal{L}_P + \lambda_G\mathcal{L}_G,
\end{align}
where $\lambda_P$ and $\lambda_G$ represent the weights for the Pseudo-label training loss and the GP guidance loss, respectively. Algorithm \ref{alg:PG-LBO} presents the pseudo-code for PG-LBO. 

\begin{algorithm}[t]
\caption{Pseudo code of PG-LBO}
\label{alg:PG-LBO}
\SetKwInOut{Input}{Input}
\Input{Data $D_L$, query budget $M$, object function $f(x)$, VAE encoder/decoder $q_\phi(x)$/$g_\theta(z)$, retrain frequency $r$, weight function $w(x)$}
\For{$1,\cdots,M/r$}{
$\mathcal{W} \gets \{w(x_i)\}_{x_i \in D_L}$ \;
$\hat{\mathcal{W}} \gets \{w(\hat{x}_j)\}_{\hat{x}_j \in D_P}$ \;
Weight $D_L$ by $\mathcal{W}$, weight $D_P$ by $\hat{\mathcal{W}}$ \;
Solve $\phi^\star, \theta^\star \gets \text{argmin}_{\phi,\theta}\mathcal{L}(\phi, \theta)$ on $D_L$, $D_P$ \;
$\{z_i \gets q_\phi(x_i)\}_{x_i \in D_L}$ \;
$D_\mathcal{Z} \gets \{(z_i, f(x_i))\}_{x_i \in D_L}$ \;
\For{$1,\cdots,r$}{
Fit GP on $D_\mathcal{Z}$ to obtain $f_\Phi(z;D_\mathcal{Z})$ \;
Optimize EI to obtain new latent point $\tilde{z}$ \;
$\tilde{x} \gets g_\theta(\tilde{z})$ \;
$D_L \gets D_L \cup \{\tilde{x}, f(\tilde{x})\}$ \; 
$D_\mathcal{Z} \gets D_\mathcal{Z} \cup \{\tilde{z}, f(\tilde{x})\}$ \;
}
Sample latent data point $\hat{z}$ \;
Filter $\hat{z}$ by thresholds $\tau$ \;
$\hat{x} \gets g_\theta(\hat{z})$ \;
$D_P \gets \{\hat{x}, \mu_{f_\Phi}(\hat{z})\}$ \;
Update $\tau$ \;
}
$x^\star \gets \text{argmax}_{x \in D_L}f(x)$ \;
\Return{$x^\star$}
\end{algorithm}

\section{Experiments}
In this section, we apply PG-LBO to three high-dimensional structured optimization tasks and compare it with several VAE-BO algorithms. For each task, we first pre-train the VAE using unlabeled data, and all algorithms start with the same pre-trained VAE as the backbone network.

\subsection{High-dimensional optimization tasks and Baselines}
\subsubsection{Topology shape fitting task:}
The goal of the task is to generate a $40 \times 40$ binary image $x$, and \textbf{maximize} the cosine similarity $\text{cos}(\mathbf{x},\mathbf{x}^\prime)=\mathbf{x} \cdot 
 {\mathbf{x}^\prime}^\top/\left \| \mathbf{x} \right \| \left \| \mathbf{x}^\prime \right \|$ between $\mathbf{x}$ and a predefined target image $\mathbf{x^\prime}$. The task involves using 10000 topology images from the dataset \cite{sosnovik2019neural} and a VAE with the latent space  dimension of 20.

\subsubsection{Expression reconstruction task:}
The expression reconstruction task aims to generate single-variable expressions $x(v)$ and \textbf{minimize} the distance to the target equation $x^\prime(v) = 1/3+v*\text{sin}(v*v)$. The objective function is a distance metric $f(x)=\max\{-7,-\int^{10}_{-10}\text{log}(1+(x(v)-x^\prime(v))^2dv\}$. Task access to 40,000 data points and use the grammar VAE from \cite{kusner2017grammar} with the latent space dimension of 25.

\subsubsection{Chemical design task:}
The task uses the ZINC250K dataset \cite{sterling2015zinc} to synthesize chemical molecules with the objective of \textbf{maximizing} the penalized water-octanol distribution coefficient (PlogP) of molecules. A Junction Tree Variational Autoencoder (JT-VAE) \cite{jin2018junction} with a latent space dimension of 56 encodes and generates efficient molecules.

\subsubsection{Baselines:}
We compare our proposed method with four VAE-BO baselines: LSBO, LBO, T-LBO and LOL-BO. LSBO \cite{gomez2018automatic} performs BO in the latent space with a fixed pre-trained VAE. LBO trains VAE on the labeled data through data weighting and periodically fine-tuning VAE. T-LBO introduces contrastive learning to LBO by additionally minimizing the triplet loss of the labeled data. LOL-BO introduces the trust region optimization strategies that aided in searching for optimal points during the BO process. We follow the setups in the existing literature and utilize the same GP surrogate and acquisition function across all baselines. The surrogate is a sparse GP \cite{titsias2009variational} with the radial basis function (RBF) kernel. The acquisition is the EI function.

\subsection{Experimental Setup and Results}
\subsubsection{Experimental setup:}
PG-LBO builds upon the foundation of LBO and uses the same VAE updating strategy and data weighting scheme during the BO process. In the training of pseudo-label data, the size of the pseudo-label dataset is maintained at half of the labeled dataset size, \textit{i.e.}, $N_P=N_L/2$. As the BO iterations progress, the accuracy of pseudo-labels improves. Therefore, we linearly increase the weight of the pseudo-label loss during VAE retraining rounds. For topology shape fitting task, $\lambda_P=\text{LinearIncrease}(0.5, 0.75)$. For expression reconstruction task and chemical design task, $\lambda_P=\text{LinearIncrease}(0.1, 0.75)$. Regarding the GP guidance loss weight, we consider the varying difficulty levels of different tasks, and accordingly, the loss weight varies. For topology shape fitting task and chemical design task, the weight $\lambda_G=1$, while for expression reconstruction task, $\lambda_G=0.1$. The momentum decay of the pseudo-label selection threshold, $\lambda=0.9$. The data sampling method employs noisy sampling, with Gaussian noise $\mathcal{N}(0, 0.1)$. 

\begin{figure*}[ht]
\centering
\includegraphics[width=0.8\textwidth]{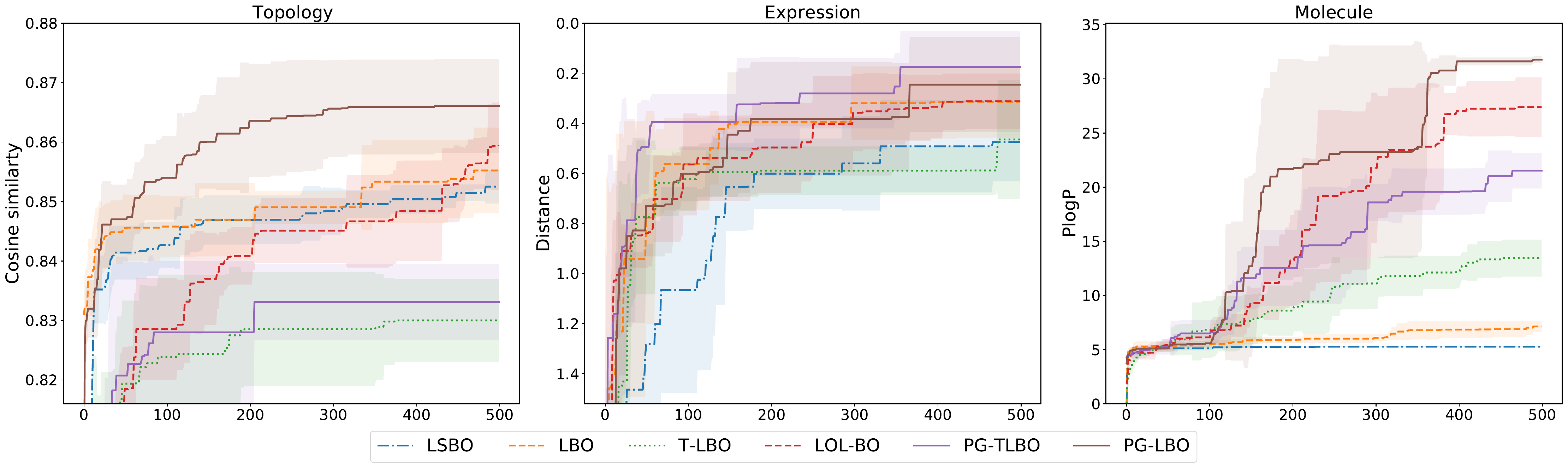} 
\caption{Comparison between PG-LBO and other VAE-BO methods in three tasks, plot the mean and the standard deviation of the best value over 5 seed over 500 evaluations. For topology task, the y-axis represents the cosine similarity between generated images and target images. For expression task, the y-axis represents the distance between generated expressions and target expressions, and for molecule task, the y-axis represents the PlogP score of generated molecules.}
\label{fig:exper-all}
\end{figure*}

\begin{table}[ht]
\centering
\renewcommand{\arraystretch}{1.0}
\resizebox{.9\columnwidth}{!} {
\begin{tabular}{c|c|c|c}
    \toprule
     \large \diagbox{Method}{Task} & \large Topology ($\uparrow$) & \large Expression ($\downarrow$) & \large Molecule ($\uparrow$) \\
    \midrule
    LSBO & 0.852$\pm$0.0028 & 0.474$\pm$0.1568 & 5.45$\pm$0.2521 \\
    LBO & 0.855$\pm$0.0072 & 0.314$\pm$0.1436 & 7.01$\pm$0.5174 \\
    T-LBO & 0.830$\pm$0.0069 & 0.464$\pm$0.2376 & 13.44$\pm$1.7110 \\
    LOL-BO & 0.859$\pm$0.0076 & 0.311 $\pm$0.1105 & 27.53$\pm$2.3927 \\
    \midrule
    \textbf{PG-TLBO (our method)} & 0.833$\pm$0.0064 & \textbf{0.175$\pm$0.1443} & 21.53$\pm$1.6503 \\
    \textbf{PG-LBO (our method)} & \textbf{0.866$\pm$0.0070} & 0.227$\pm$0.1864 & \textbf{31.77$\pm$0.3287}\\
    \bottomrule
\end{tabular}
}
\caption{Mean and standard deviation of best values found by baseline methods and PG-LBO on three tasks. For Topology task, the value represents cosine similarity, for Expression task, the value represents distance, and for Molecule task, the value represents PlogP score. $\uparrow$ means the higher the better, $\downarrow$ means the lower the better.}
\label{tab:result}
\end{table}

\subsubsection{Results:}
As shown in Figure \ref{fig:exper-all}, PG-LBO consistently outperforms all baselines by the end of optimization. Table \ref{tab:result} shows more details. We notice that the improvement of PG-LBO over LBO is not substantial in the topology task. This might be attributed to the fact that LBO already performs well in this task, and due to the inherent difficulty of the problem, achieving significant enhancements becomes challenging. In expression task and molecule task, PG-LBO shows a notable improvement over LBO. The experimental results of T-LBO are obtained by replicating the code and parameters provided in T-LBO paper. However, the results in all three tasks significantly deviate from those reported in T-LBO paper. To address this, we conducted additional experiments by applying our proposed method to T-LBO, named PG-TLBO. Notably, the performance of model is improved after incorporating our method. In expression task, PG-TLBO (0.175$\pm$0.1443) even outperforms PG-LBO (0.227$\pm$0.1864), demonstrating the effectiveness of our proposed method. We also conducted comparative experiments on a smaller scale labeled dataset, our method consistently maintained superior performance over all baseline methods. The detailed discussion and comparison are in Appendix C.

\subsection{Ablation Studies}
Unless otherwise specified, we conduct ablation studies on the topology shape fitting task. In the paper, we primarily present ablation experiments on the modules and key designs of the proposed methods. Ablation experiments for the hyperparameters are deferred to Appendix D.

\begin{figure*}[ht]
    \centering
    \hfill
    \begin{subfigure}[b]{0.27\textwidth}
        \includegraphics[width=\textwidth]{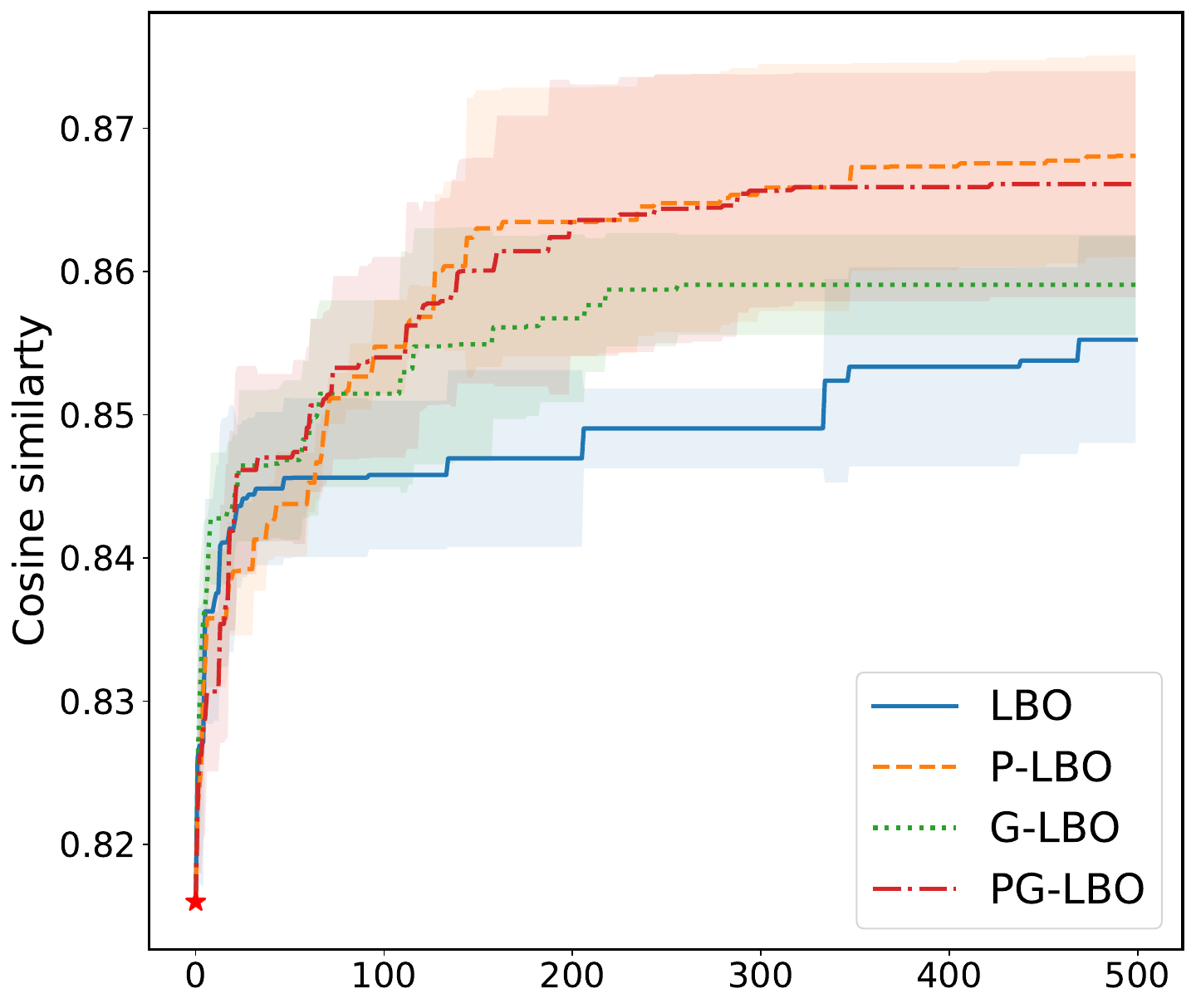}
        \caption{Impact of method module.}
        \label{fig:pglbo}
    \end{subfigure}
    \hfill
    \begin{subfigure}[b]{0.27\textwidth}
        \includegraphics[width=\textwidth]{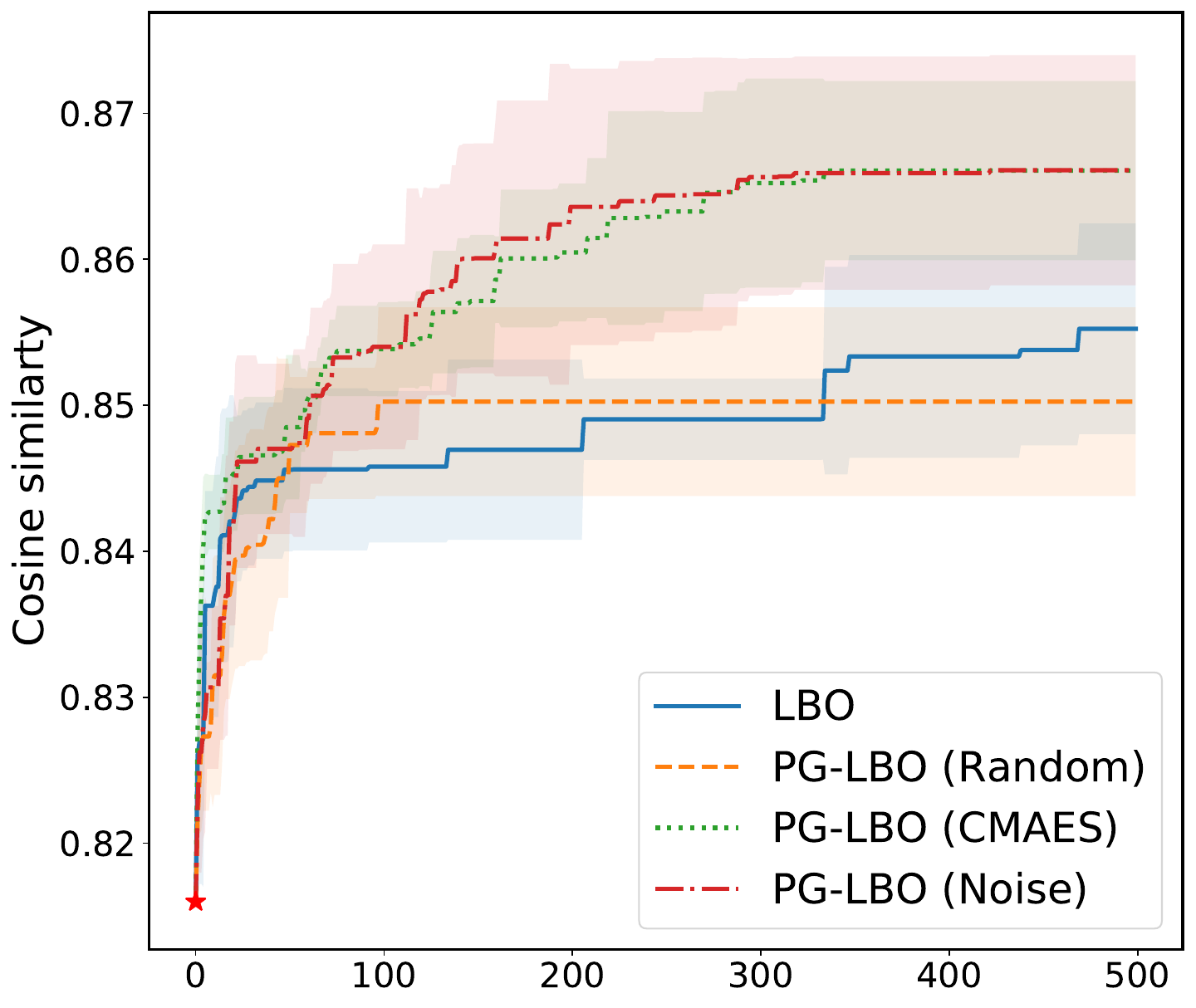}
        \caption{Impact of sampling method.}
        \label{fig:sampling}
    \end{subfigure}
    \hfill
    \begin{subfigure}[b]{0.27\textwidth}
        \includegraphics[width=\textwidth]{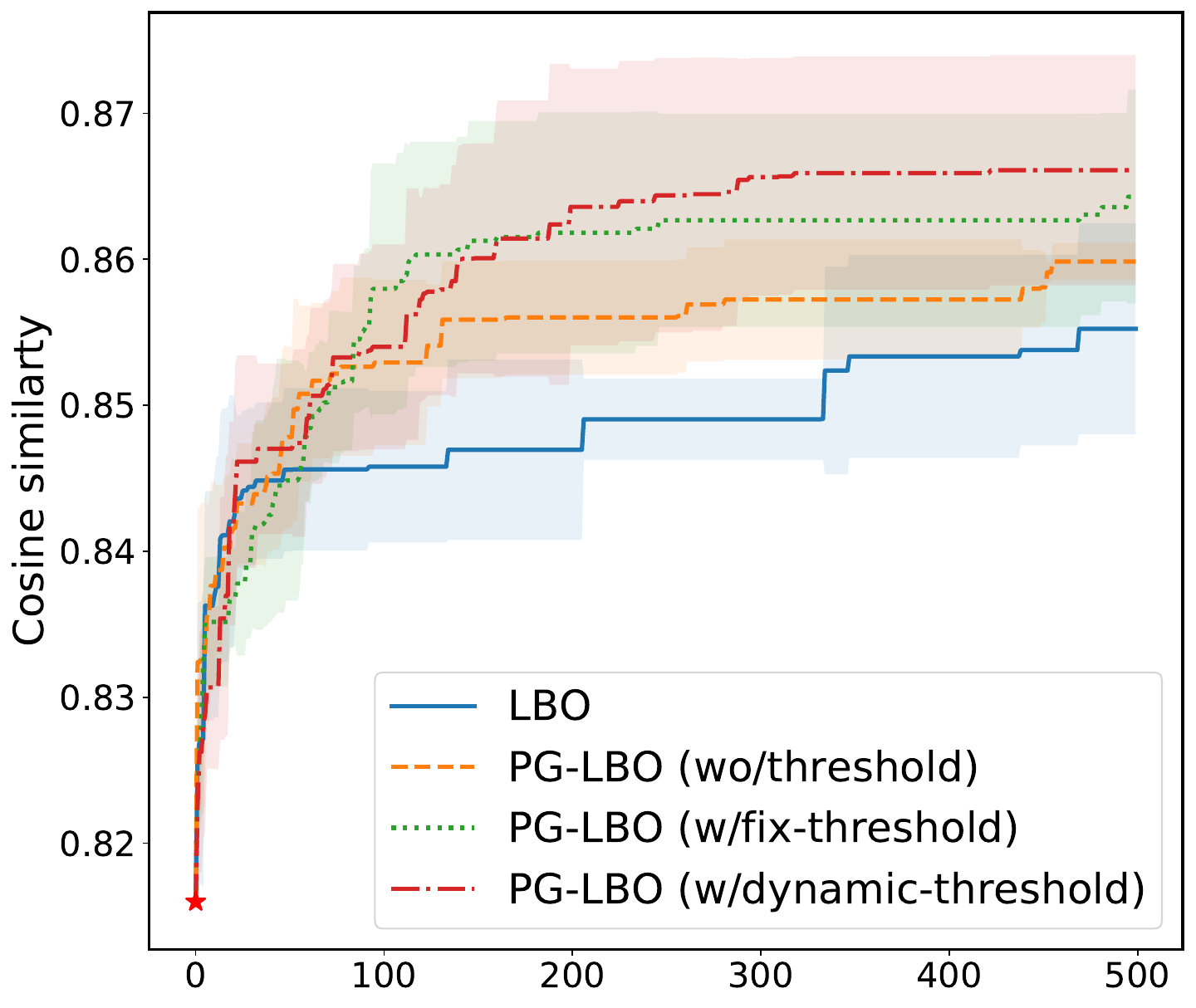}
        \caption{Impact of pseudo-label selection.}
        \label{fig:threshold}
    \end{subfigure}
    \caption{Ablation studies result, plot the mean and the standard deviation of the best value over 5 seed over 500 evaluations. The y-axis represents the cosine similarity.}
    \label{fig:ablation-result}
\end{figure*}

\subsubsection{Individual effectiveness of Pseudo-labeled training and GP guidance.}
\begin{figure}[ht]
\centering
\includegraphics[width=0.8\columnwidth]{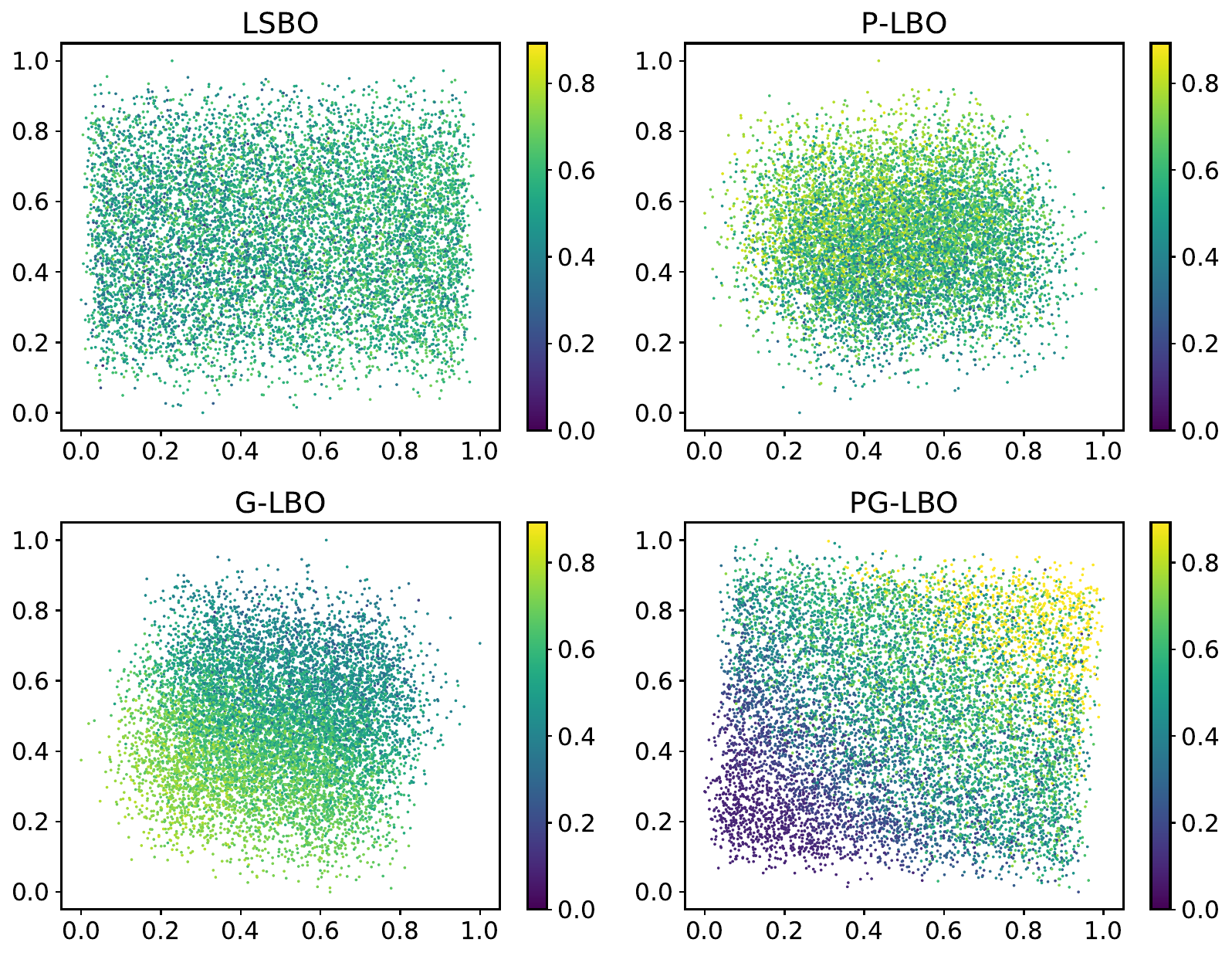}
\caption{Two-dimensional PCA analysis on the VAE latent space for the topology shape fitting task.}
\label{fig:latent-space}
\end{figure}

To investigate the impact of the two main components of our method, pseudo-label training and GP guidance, on model performance, we conduct ablation experiments and attempt to visualize the VAE latent space. \textbf{Experimental Setup:} We compare three scenarios: using only pseudo-label training, using only GP guidance, and combining both techniques, denoted as P-LBO, G-LBO, and PG-LBO respectively. Other settings remain consistent. \textbf{Result:} As shown in Figure \ref{fig:pglbo}, Pseudo-label training (P-LBO) and GP guidance (G-LBO) both contribute to performance improvements in the model, and for the topology task, Pseudo-label training provides more significant improvement to the model's performance. This might be attributed to the fact that for topology tasks, the GP model has already fitted quite well, thus the improvement from GP guidance is limited. We conduct a visual analysis to further investigate the impact of pseudo-label training and GP guidance on the VAE latent space. We randomly sample 10,000 points from the trained VAE latent space and compute their corresponding true label values. We then perform PCA dimensionality reduction for visualization. The results are shown in Figure \ref{fig:latent-space}, the two axes are the principal components selected from the PCA analysis. The color bar represents the values of the selected property (cosine similarity). LSBO represents an unsupervised trained VAE. We can observe that the distribution of points in the latent space of P-LBO, G-LBO, and PG-LBO exhibits greater discriminative characteristics.

\subsubsection{The impact of data sampling on model performance.}
The quality of sampling unlabeled data significantly impacts the model's performance. We conduct ablation experiments to compare different sampling methods. \textbf{Experimental setup:} We compare three sampling methods: random sampling, heuristic sampling, and noisy sampling, denoted as PG-LBO (Random), PG-LBO (CMAES), and PG-LBO (Noise) respectively. In the Methodology section, we provide a detailed description of the implementation process for heuristic sampling and noisy sampling. We employ the Covariance Matrix Adaptation Evolution Strategy (CMA-ES) algorithm for heuristic sampling as the optimization algorithm. We select 100 high-scoring points from the existing labeled dataset as the initial means for the CMA-ES algorithm. The initial standard deviation is set to 0.25, and we perform 100 iterations, using the points generated in each iteration as sampling points. For noisy sampling, we sample noise from a Gaussian distribution $\mathcal{N}(0, 0.1)$. Similar to heuristic sampling, we select 100 high-scoring points from the existing labeled dataset as seed points, add random Gaussian noise to these seed points, and use the newly obtained points as sampling points. Aside from the sampling method, other parameter settings remain consistent. \textbf{Result:} As shown in Figure \ref{fig:sampling}, the performance of random sampling is the worst, even lower than the baseline (LBO). This indicates that the effectiveness of the pseudo-label training is highly dependent on the quality of the unlabeled data points sampled. If the sampled points are of poor quality, pseudo-label training might even degrade the performance compared to the baseline. Heuristic sampling and noisy sampling exhibit comparable performance. However, since heuristic sampling involves executing an additional optimization algorithm, it introduces additional computational overhead. As a result, we ultimately choose simple and effective noisy sampling as the primary sampling method.

\subsubsection{The role of pseudo-label selection.}
To improve the quality of pseudo-labels, we select pseudo-label data based on a dynamic threshold derived from uncertainty. We perform ablation experiments to investigate the impact of pseudo-label thresholds on model performance. \textbf{Experimental setup:} We compare three scenarios: using a fixed threshold, using a dynamic threshold, and using no threshold, denoted as PG-LBO (w/fix-threshold), PG-LBO (w/dynamic-threshold), and PG-LBO (wo/threshold) respectively. The fixed threshold is set to 0.0015, and the momentum decay $\lambda$ for the dynamic threshold scheme is set to 0.9. Other experimental settings remain consistent. \textbf{Results:} As shown in Figure \ref{fig:threshold}, after threshold filtering, there is an improvement in model performance, indicating that enhancing the accuracy of pseudo-labels contributes to better model performance. The dynamic threshold filtering scheme performs better than the fixed threshold scheme, suggesting that dynamically adjusting the threshold based on the model's learning state is beneficial. Moreover, the fixed threshold scheme requires task-specific prior knowledge to determine the threshold, whereas the dynamic threshold scheme does not.

\section{Conclusion}
In this paper, we propose a novel method to utilize unlabeled data through pseudo-labeling techniques in semi-supervised learning. Our method employs pseudo-labels to reveal discriminative information about the optimization objective value hidden in the unlabeled data and uses this information to enhance the construction of the VAE latent space. Furthermore, drawing inspiration from deep kernel learning, we integrate the VAE encoder and the GP model of the BO as a unified process and directly incorporate the goal of improving GP prediction accuracy into the training of the VAE, providing a new way of exploiting labeled data. By combining our method with the weighted retraining proposed by LBO, we obtain a novel BO algorithm named PG-LBO. Extensive experiments on multiple task datasets demonstrate the effectiveness of our method in enhancing the performance of existing VAE-BO methods.

\bibliography{aaai24}
\clearpage

\setcounter{figure}{0} 
\setcounter{table}{0}
\renewcommand\thefigure{\Alph{figure}} 
\renewcommand\thetable{\Alph{table}}
\appendix
\subsection{A Detailed Explanation of Sampling Method}
\begin{figure}[h]
\centering
\includegraphics[width=\columnwidth]{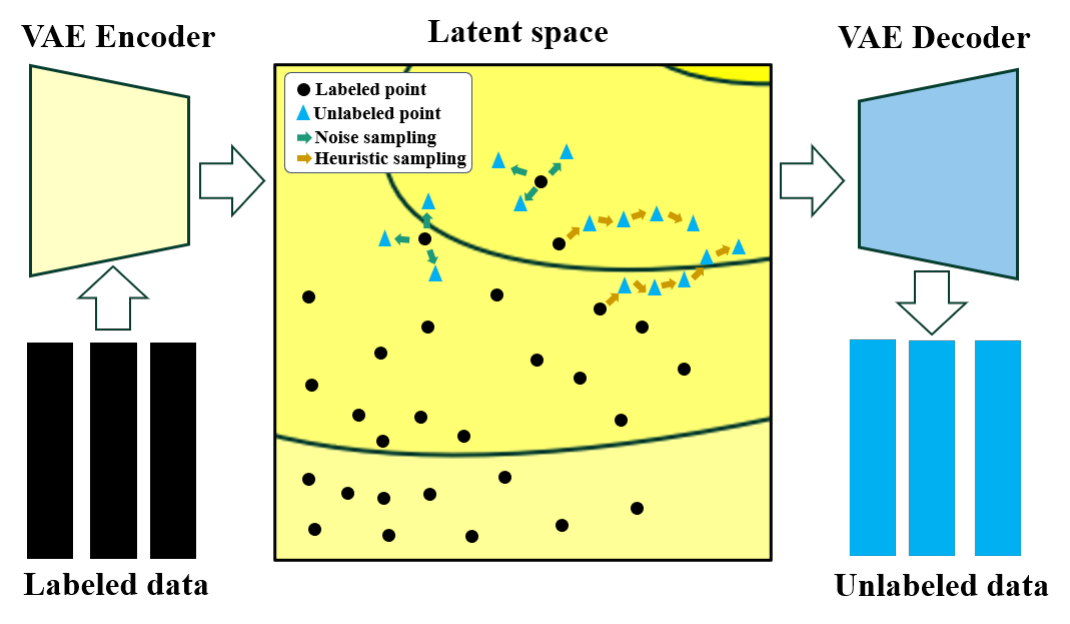}
\caption{Explanation of sampling method.}
\label{fig:sampling-appendix}
\end{figure}

We present two straightforward yet effective sampling methods designed to procure high-quality points. Illustrated in Figure \ref{fig:sampling-appendix}, the noise sampling entails the introduction of Gaussian noise to the latent representation of existing points, yielding sample points in close proximity. This method is akin to sampling in the vicinity of existing data points. Conversely, the heuristic sampling initiates from existing points and employs a heuristic optimization algorithm aimed at maximizing the posterior mean of the Gaussian Process (GP) model. The points generated during the optimization iterations are chosen as sample points. The entire sampling process takes place in the latent space, and upon obtaining the sample points, they are seamlessly mapped back to the original input space through a Variational Autoencoder (VAE) decoder, serving as the final pseudo-labeled data.

\subsection{B The Explanation on Threshold Setting}
\begin{figure}[h]
\centering
\includegraphics[width=\columnwidth]{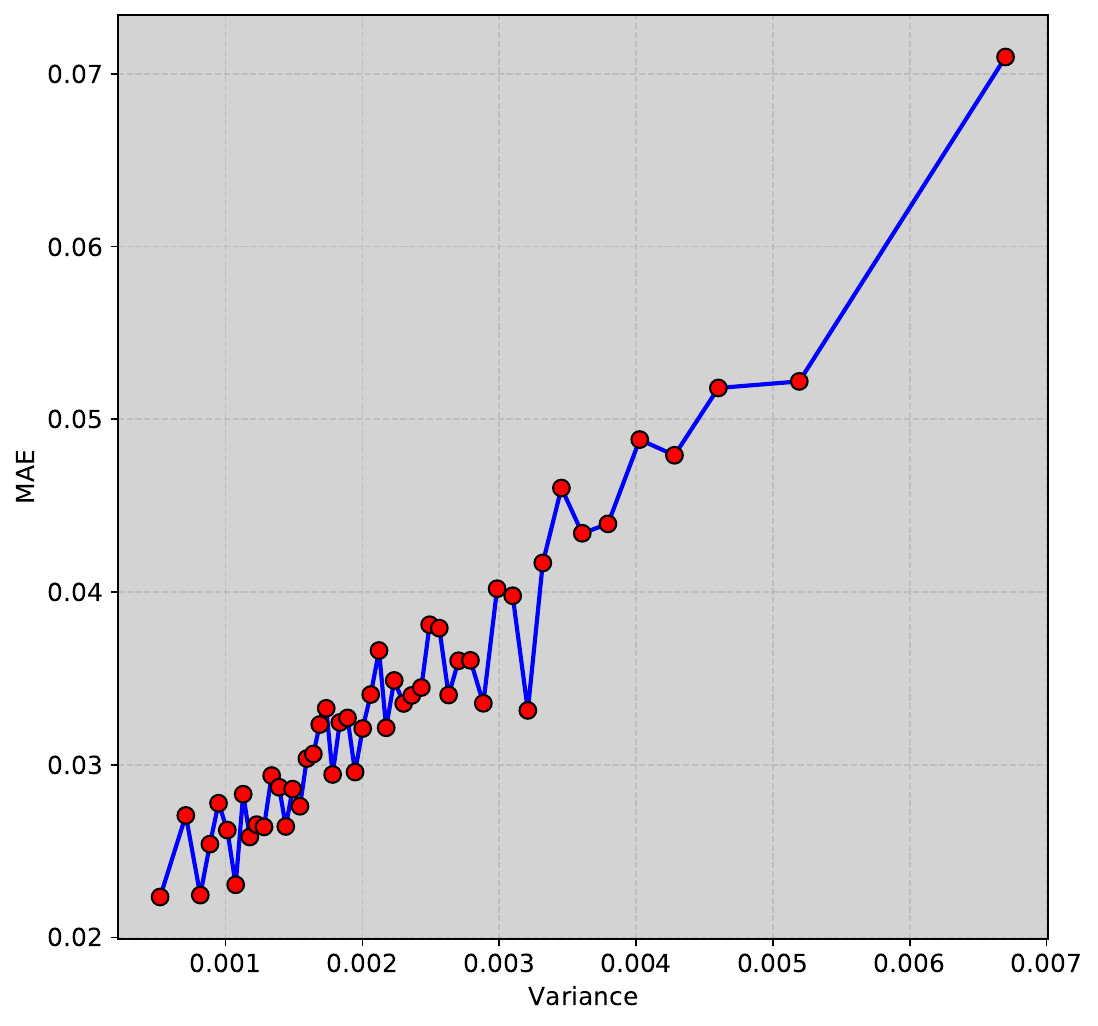}
\caption{Explanation of threshold setting.}
\label{fig:thredshold-appendix}
\end{figure}

To elucidate our rationale for choosing uncertainty as the threshold for pseudo-label selection, we conducted supplementary analysis to underscore our motivation. In the topology shape fitting task, we sampled 5000 data points and recorded the pseudo-labels (posterior means) and posterior variances provided by the GP model for each point. Subsequently, we computed the Mean Absolute Error (MAE) between the pseudo-labels and the corresponding ground-truth labels for each point, serving as a representation of pseudo-label errors. Organizing all data points in ascending order of their variances, we grouped them in sets of 100 points each. The variance mean for each group was then computed for the $x$-axis, and the corresponding MAE mean for the $y$-axis. These values were plotted as a line chart. As illustrated in Figure \ref{fig:thredshold-appendix}, a significant positive correlation between MAE and variance emerged, smaller variances correspond to smaller pseudo-label errors, substantiating the rationality of using posterior variance as a selection threshold.

\subsection{C Experiment on a Small-Scale Labeled Dataset}
To investigate the performance of our method on datasets with fewer labeled samples, we conducted a comparative experiment on a small-scale labeled dataset. In this experiment, we utilized less than 1\% of the labeled data (Topology [100/10k], Expression [100/40k], Molecule [1000/250k]) and set the size of the pseudo-labeled dataset to ten times that of the labeled dataset ($N_P = 10 N_L$), while keeping all hyperparameters unchanged.

\subsubsection{Results:}
\begin{figure*}[ht]
\centering
\includegraphics[width=\textwidth]{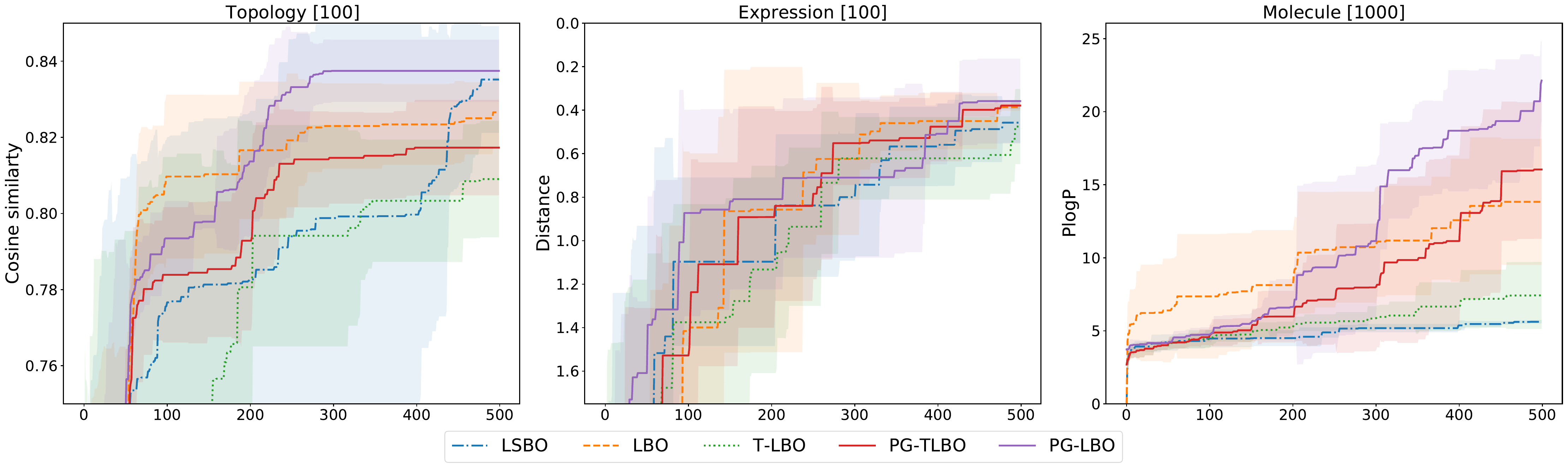} 
\caption{Comparative experiment results on smaller scale labeled dataset.}
\label{fig:semi-exper-all}
\end{figure*}

As illustrated in Figure \ref{fig:semi-exper-all}, the performance of all methods exhibited a decline when dealing with a smaller labeled dataset. Nevertheless, PG-LBO consistently outperformed all baseline models by the conclusion of the optimization process. Additional insights and detailed results are available in Table \ref{tab:smaller}. Notably, in the context of the chemical design task, we observed suboptimal performance using the official implementations based on Tensorflow \cite{tensorflow2015-whitepaper} and GPflow \cite{GPflow2017} under limited data conditions. Consequently, for the chemical molecule synthesis task, we made modifications to the implementation, transitioning to PyTorch \cite{Paszke_PyTorch_An_Imperative_2019} and Botorch \cite{balandat2020botorch}. It is crucial to emphasize that all comparative methods were implemented within the same framework, ensuring uniformity and consistency across the evaluations.

\begin{table}[h]
\centering
\renewcommand{\arraystretch}{1.0}
\resizebox{\columnwidth}{!} {
\begin{tabular}{c|c|c|c}
    \toprule
      \diagbox{Method}{Task} &  Topology [100] ($\uparrow$) &  Expression [100] ($\downarrow$) &  Molecule [1000]* ($\uparrow$) \\
    \midrule
    LSBO &  0.835$\pm$0.0140 &  0.457$\pm$0.0927 &  5.61$\pm$0.1198 \\
     LBO &  0.826$\pm$0.0092 &  0.386$\pm$0.0160 &  13.83$\pm$4.3147 \\
     T-LBO &  0.809$\pm$0.0152 &  0.475$\pm$0.1727 &  7.42$\pm$2.2970 \\
    \midrule
    \textbf{PG-TLBO (our method)} &  0.817$\pm$0.0125 &  0.379$\pm$0.0147 &  16.04$\pm$4.7419 \\
    \textbf{PG-LBO (our method)} &  \textbf{0.837$\pm$0.0081} &  \textbf{0.358$\pm$0.1950} &  \textbf{23.23$\pm$2.9129}\\
    \bottomrule
\end{tabular}
}
\caption{Result on small scale labeled dataset. *: Implemented using PyTorch and Botorch.}
\label{tab:smaller}
\end{table}

\subsection{D The Ablation Studies for Hyperparameters}
\begin{figure*}[ht]
\centering
\includegraphics[width=\textwidth]{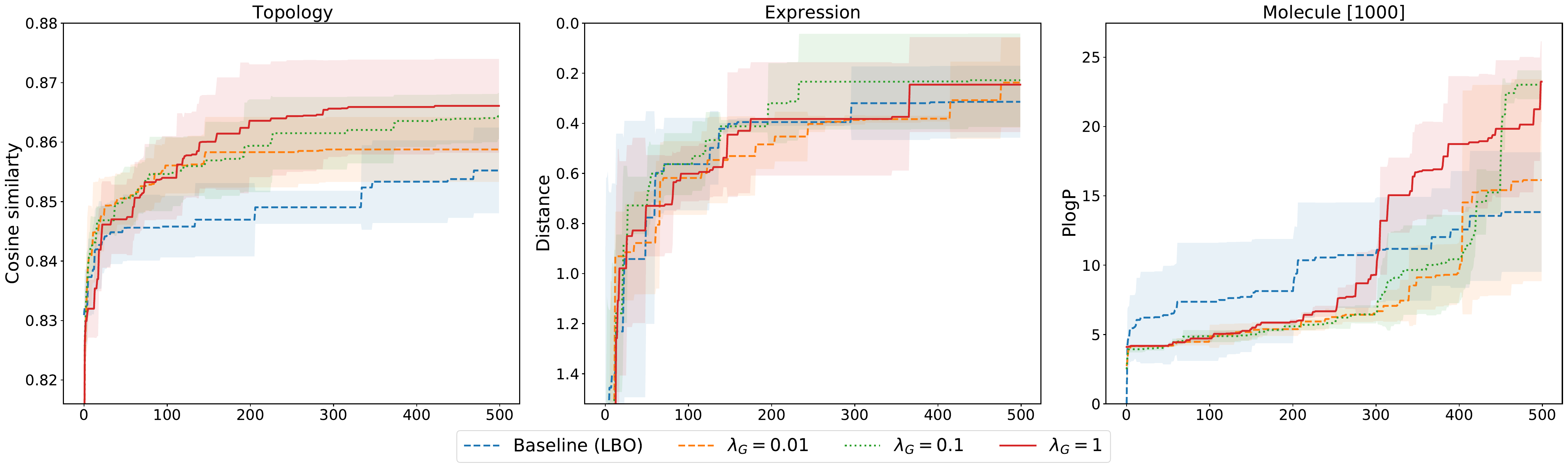} 
\caption{Ablation experiment results for $\lambda_G$.}
\label{fig:exper-lg}
\end{figure*}

\begin{figure*}[ht]
\centering
\includegraphics[width=\textwidth]{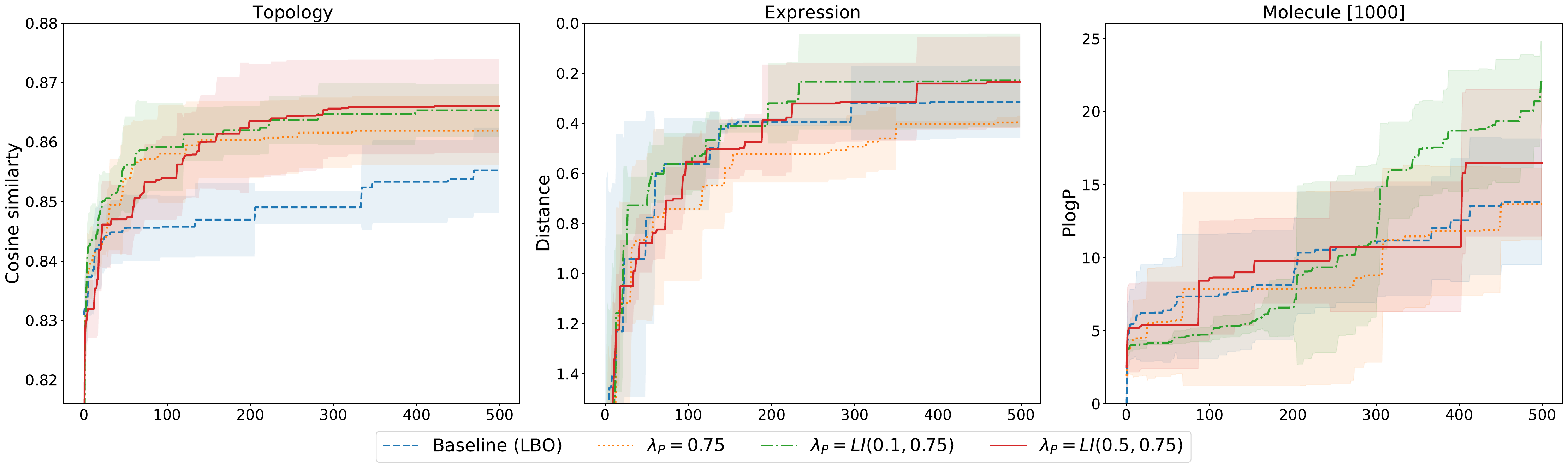} 
\caption{Ablation experiment results for $\lambda_P$.}
\label{fig:exper-lp}
\end{figure*}

To examine the influence of hyperparameters on experimental outcomes and provide insights into the rationale behind specific hyperparameter choices, we conducted sensitivity experiments outlined in Table \ref{tab:hyperparameters}. \textbf{Experimental Setup:} Our investigation centered on two crucial hyperparameters: $\lambda_P$ with values \{0.01, 0.1, 1\}, and $\lambda_G$ with values \{0.75, LinearIncrease(0.5, 0.75), LinearIncrease(0.1, 0.75)\}. Throughout the experiments, all other configurations remained consistent with previous comparative studies, except for the hyperparameters under scrutiny. It is noteworthy that the chemical design task exhibited analogous trends across various methods under both smaller labeled dataset and full labeled dataset settings. Considering limitations in terms of time and computational costs, we specifically conducted hyperparameter ablation experiments for the chemical design task with a smaller labeled data setting. \textbf{Results:} The detailed outcomes are presented in Table \ref{tab:hyperparameters}, accompanied by corresponding visual representations in Figures \ref{fig:exper-lg} and \ref{fig:exper-lp}. Notably, the final optimization results consistently outperformed the baseline LBO across all values of $\lambda_G$. Different values of $\lambda_G$ exhibited minimal impact on the model's ultimate performance. Regarding $\lambda_P$, the superiority of employing a simple linearly increasing weight over a fixed weight was evident. As the BO process progresses, the accuracy of pseudo-labels gradually improves. The linearly increasing weight better leverages this trend, avoiding excessive confirmation bias in the early stages. Furthermore, optimal hyperparameter values varied across different tasks. Given substantial differences in input data and network structures for various tasks, such variability is reasonable. Ultimately, for each task, we selected the optimal hyperparameter settings.

\begin{table}[h]
\centering
\renewcommand{\arraystretch}{1.0}
\resizebox{\columnwidth}{!} {
\begin{tabular}{c|c|c|c}
    \toprule
      \diagbox{Hyperparameter}{Task}  &  Topology ($\uparrow$) &  Expression ($\downarrow$) &  Molecule [1000]*($\uparrow$) \\
    \midrule
     \textbf{$\lambda_G=1$} &  \textbf{0.866$\pm$0.0070} &  0.237 $\pm$0.1795 &  \textbf{23.23$\pm$2.9129} \\
     \textbf{$\lambda_G=0.1$} &  0.865$\pm$0.0044 &  \textbf{0.227$\pm$0.1864} &  23.01$\pm$1.0427 \\
     \textbf{$\lambda_G=0.01$} &  0.863$\pm$0.0041 &  0.245$\pm$0.1895 &  16.13$\pm$7.2849 \\
    \midrule
     \textbf{$\lambda_P=0.75$} &  0.863$\pm$0.0041 &  0.392$\pm$0.0235 &  13.68$\pm$2.4674 \\
     \textbf{$\lambda_P=LinearIncrease(0.5,0.75)$} &  \textbf{0.866$\pm$0.0070} &  0.235$\pm$0.1815 &  16.50$\pm$5.0343\\
     \textbf{$\lambda_P=LinearIncrease(0.1,0.75)$} &  0.865$\pm$0.0044 &  \textbf{0.227$\pm$0.1864} &  \textbf{23.23$\pm$2.9129} \\    
    \bottomrule
\end{tabular}
}
\caption{Sensitivity experiments result for $\lambda_G$ and $\lambda_P$. *: Implemented using PyTorch and Botorch.}
\label{tab:hyperparameters}
\end{table}

\end{document}